\documentclass[runningheads]{llncs}

\usepackage[paperheight=235mm, paperwidth=155mm,textwidth=12.2cm,textheight=19.3cm]{geometry}
\usepackage[T1]{fontenc}
\usepackage{microtype}

\usepackage{enumitem}
\usepackage{graphicx}
\usepackage{wrapfig}
\usepackage{lipsum}
\usepackage{academicons}
\usepackage{cite}
\usepackage{caption, subcaption}
\usepackage{booktabs}

\usepackage{xcolor}

\makeatletter\let\proof\@undefined\let\endproof\@undefined\makeatother

\usepackage{amsmath,amssymb,amsthm}
\usepackage{mathrsfs}
\usepackage{physics}

\usepackage{numprint}
\npthousandsep{\,}

\usepackage{algorithm}
\usepackage[noend]{algpseudocode}

\usepackage{booktabs}

\usepackage{tikz}
\usetikzlibrary{shapes,positioning,arrows,calc,automata,matrix}

\usepackage{amsmath}

\usepackage{bm}

\newcommand{\diff}{\mathop{}\!d}

\usepackage{authblk}
\title{Towards Invertible Semantic-Preserving Embeddings of Logical Formulae}
\author{Gaia Saveri\inst{1,2} \and
Luca Bortolussi\inst{2}}
\date{}
\institute{Department of Computer Science, University of Pisa, Italy\and
Department of Mathematics and Geoscience, University of Trieste, Italy}

\begin{document}

\maketitle

\vspace*{-1em}

\begin{abstract}
  Logic is the main formal language to perform automated reasoning, and it is further a human-interpretable language, at least for small formulae. Learning and optimising logic requirements and rules has always been an important problem in Artificial Intelligence. State of the art Machine Learning (ML) approaches are mostly based on  gradient descent optimisation in continuous spaces, while learning logic is framed in the discrete syntactic space of formulae. Using continuous optimisation to learn logic properties is a challenging problem, requiring to embed formulae in a continuous space in a meaningful way, i.e.  preserving the semantics. Approaches like \cite{stlkernel} are able to construct effective semantic-preserving embeddings via kernel methods (for linear temporal logic), but the map they define is not invertible. In this work we address this problem, learning how to invert such an embedding leveraging deep architectures based on the Graph Variational Autoencoder framework. We propose a novel model specifically designed for this setting, justifying our design choices through an extensive experimental evaluation. Reported results in the context of propositional logic are promising, and several challenges regarding learning invertible embeddings of formulae are highlighted and addressed.
\end{abstract}

\vspace*{-2em}

\section{Introduction}\label{sec:intro}

Logic, in its many variants, is arguably one of the most prominent languages used to represent knowledge, specify requirements and explanations for complex systems, and reason in a human-comprehensible way \cite{ai}. On the other hand, Graph Neural Networks (GNN, \cite{gnn-original}) have reached state-of-the-art in relational learning, providing meaningful inductive biases such as permutation invariance and sparsity awareness \cite{gnn-survey, inductive-bias}.  Lately, much attention has been put on combining symbolic knowledge representation and neural computations as performed by GNNs, towards solving combinatorial and logical reasoning tasks  \cite{gnn-nesy, combopt}.
In this context, leveraging GNNs to learn real-valued representations of logic formulae would open the door to the use of gradient-based optimization techniques in the space of formulae, hence moving requirement mining from a discrete to a continuous search problem. A key desiderata of such a model would be that of semantic consistency, i.e. formulae which are semantically similar should be mapped to vectors which are close in the latent space. Current state-of-the-art in this field, for Linear Temporal Logic, is based on kernel methods, hence it learns a non-invertible function of formulae to their embeddings \cite{stlkernel}. We address this drawback proposing a model based on the Graph Variational Autoencoder framework (GVAE, \cite{gvae, gvae-survey}), whose objective is to construct an invertible mapping from the discrete syntactic space of logic formulae to a continuous space where semantic similarity is preserved. 

\paragraph{Our Contribution} Moving from the rationale of GVAE models for Direct Acyclic Graphs (DAG, \cite{dvae}), we formulate a model able to encode the syntactic tree of a logic formula into a continuous discriminative representation. Towards the goal of making these embeddings semantic-preserving, we propose a principled strategy  to enrich the learned latent space with semantic information. We show results in the context of propositional logic, which - despite its grammatical simplicity - already poses challenges arising from the intrinsic symmetry of propositions and from the syntax-semantic interplay, which we address via a series of justified design choices. Experimental results are promising and should be taken as a proof-of-concept towards extending this methodology to richer logic formalisms. 

\paragraph{Related Work} Learning autoencoder models for DAGs was originally addressed via sequence models \cite{dag-2, dag-1, dag-3} , then casted into a graph-related task in \cite{dvae, dagnn, graph-gen-1}. A significant body of work in this field is dedicated to molecular generation \cite{mol-gen-1, mol-gen-2, mol-gen-3}.  Learning generative models with either syntactic or semantic constraints is instead tackled in \cite{constrained-2, constrained-1}, respectively. Constructing a semantic similarity measure for logic formulae is instead in \cite{stlkernel}.

\section{Background}\label{sec:background}
As mentioned in Section \ref{sec:intro}, in this work we restrict the investigation to the setting of propositional logic. Denoting by $\{x_i\}_{i=1}^n$ a set of propositional variables, where each $x_i$ is a binary variable that can evaluate either to true ($1$) or false ($0$), the set of propositional formulae (hereafter referred to also as propositions or requirements) is defined recursively by the syntax $\varphi:=1\mid x_i \mid \neg \varphi \mid \varphi_1 \wedge \varphi_2 \mid \varphi_1 \vee \varphi_2$. Semantics is defined recursively following the usual meaning of Boolean operators (see Appendix \ref{subapp:logic} for details).  

A convenient way to represent propositions in this context is by means of their abstract syntax tree (AST), i.e. a tree representation of the syntactic structure of the formula where internal nodes are the Boolean operators and the leaves are the propositional variables. Since Boolean operators are at most binary, the AST arising from propositional formulae are binary trees. This representation opens the doors to the application of Graph Neural Network (GNN, \cite{gnn-original}) models, i.e. deep learning models natively able to handle graph-structured inputs. 

In this work we wish to learn both an encoder and a decoder for mapping inputs $\textbf{x}$ to and from continuous vectors $\textbf{z}$. The Variational Autorencoder (VAE, \cite{vae}) framework is designed to learn simultaneously two parametric functions: a probabilistic generation network (decoder) $p_{\theta}(\textbf{x}|\textbf{z})$ and an approximated posterior distribution (encoder) $q_{\phi}(\textbf{z}|\textbf{x})$, by maximizing the evidence lower bound:
\begin{equation}
    \mathcal{L}(\phi, \theta; \textbf{x}) = \mathbb{E}_{\textbf{z}\sim q_{\phi}(\textbf{z}|\textbf{x})} [\log p_{\theta}(\textbf{x}|\textbf{z})] - KL[q_{\phi}(\textbf{z}|\textbf{x})||p(\textbf{z})]
    \label{eq:vae-loss}
\end{equation}
where $KL[q(\cdot)||p(\cdot)]$ denotes the Kullback-Leibler divergence between $q(\cdot)$ and $p(\cdot)$ and $p(\textbf{z})$ the prior distribution over latent variables $\textbf{z}$.

The VAE model has been generalized to graph structured data by employing a GNN as encoder (and possibly decoder) model \cite{gvae}. Deep generative models for graphs are partitioned into two classes (borrowing the terminology from \cite{graph-gen-survey-1, graph-gen-survey-2}), following the way in which nodes and edges of the graph are produced: \textit{sequential generating} if nodes and edges are processed following a predefined order on them, one at a time; \textit{one-shot generating} if the adjacency matrix of the graph is generated all at once.

If auxiliary information $\textbf{y}$ (such as category labels or semantic context) about the input $\textbf{x}$ should be incorporated into the learning process, frameworks like the Conditional Variational Autoencoder (CVAE, \cite{cvae, cond-vae}) allow to control the generative process by imposing a condition on both the encoder (in this context defined as $q_{\phi}(\textbf{z}|\textbf{y}, \textbf{x})$), the prior ($p_{\psi}(\textbf{z}|\textbf{y)}$, which is parametric) and the decoder ($p_{\theta}(\textbf{x}|\textbf{y}, \textbf{z})$). The network is trained by optimizing the following conditional marginal log-likelihood (notation is the same of Equation \ref{eq:vae-loss}):
\begin{equation}
    \mathcal{L}(\phi, \theta, \psi; \textbf{x}, \textbf{y}) = \mathbb{E}_{\textbf{z}\sim q_{\phi}(\textbf{z}|\textbf{y}, \textbf{x})} [\log p_{\theta}(\textbf{x}|\textbf{y}, \textbf{z})] - KL[q_{\phi}(\textbf{z}|\textbf{y}, \textbf{x})||p_{\psi}(\textbf{z}|\textbf{y})]
    \label{eq:cvae-loss}
\end{equation}

While for euclidean input formats the vector $\textbf{y}$ is commonly concatenated to the input $\textbf{x}$ (resp. the latent $\textbf{z}$) before the computation of the encoding (resp. decoding) functions, in the context of graph generative models several possibilities exist: in cases in which there is additional information for all the nodes of the graph, it is concatenated to the initial node states \cite{cond-gvae-nodes}; when $\textbf{y}$ is a property of the whole graph, it can be either concatenated to the latent vector $\textbf{z}$ before the decoding process \cite{cond-gvae-z} or incorporated into the GNN decoding architecture itself \cite{cond-gvae-mpnn}.

\subsection{Kernel-based Logic Embeddings} \label{sec:propkernel}
Learning embeddings of logic formulae preserving semantic similarity is addressed in \cite{stlkernel} by means of the kernel trick. In more detail, the problem of finding a similarity measure between Signal Temporal Logic (STL) formulae is tackled by defining a suitable kernel function (and a corresponding learning algorithm for it) mapping formulae to a subspace of the continuous Hilbert space $L^2$, resulting in a high value for semantically similar formulae (and a low value otherwise; we refer to Appendix \ref{app:kernel} for more details). Being the Boolean operators a subset of the operators of STL, we can restrict the definition of the kernel in \cite{stlkernel} to the context of propositional logic. Considering propositions with at most $n$ variables, the set $\mathcal{T}$ of the $2^n$ possible variable configurations and denoting $\varphi(\tau)=1$ (resp. $\varphi(\tau)=-1$) if the formula $\varphi$ evaluated on $\tau\in \mathcal{T}$ returns true (resp. false), then we formulate a \textit{Boolean} kernel (by restricting the definition of Boolean \textit{STL} kernel in \cite{stlkernel}) between formulae $\varphi, \psi$ as:
\begin{equation}
    k(\varphi, \psi) = \int_{\tau\in\mathcal{T}} \varphi(\tau) \cdot \psi(\tau) \diff \tau
    \label{eq:boolean-kernel}
\end{equation}
and compute it using Monte Carlo approximation putting a uniform probability measure over the space of configurations $\mathcal{T}$.
Experiments confirm that the Boolean kernel of Equation \ref{eq:boolean-kernel} maps propositional formulae in a semantic-preserving continuous space, as reported in Appendix \ref{app:kernel}, where we also show its correlation with Jaccard similarity coefficient among propositional formulae. 
Moreover, reducing the dimensionality of the embedding space using Principal Component Analysis (PCA) shows that it is sensible to consider an embedding space of dimension much lower than the number of formulae used to construct the kernel of Equation \ref{eq:boolean-kernel}. We defer to Appendix \ref{app:kernel} for experimental results on dimensionality reduction. 

\section{Logic Embeddings by Graph Neural Networks}\label{sec:method}
The main objective of this work is to investigate algorithms for computing invertible embeddings of propositional logic formulae, starting from their syntax and exploring methods for including semantic information, towards the goal of learning a latent space characterising the semantic of the logic. As every formula can be represented by its AST without any loss of information (see Section \ref{sec:background} and Appendix \ref{app:background}), architectures based on the GNN framework provide a sensible inductive bias for our setting. Moreover, as we require to learn both an encoder and a decoder for formulae, we find it meaningful to deploy a Graph Variational Autoencoder (GVAE) model. 
Formally, given a propositional formula $\varphi$ sampled according to the algorithm detailed in Appendix \ref{subapp:data}, we design the input to our model to be the rooted tree arising from its AST, i.e. inputs are graphs $G=(V, E)$ with set of vertices $V$ corresponding to operators and variables (taken with their multiplicity) appearing in $\varphi$ and set of directed edges $E\subseteq V\times V$ such that $(v_i, v_j)\in E$ if $v_j$ is a main subformula of $v_i$ in $\varphi$. 
Although normal forms exists for propositional logic, we decided not to adopt any of them for mainly two reasons: firstly, rewriting formulae in normal form usually involves increasing the number of terms (namely every formula in $n$ variables written either in conjunctive or disjunctive normal form can have up to $2^n$ terms) and this can yield to scalability issues when the formula is used as input for a GNN; secondly, in this work we are using propositional logic as a proof of concept for models that we wish to deploy for temporal logics, which in general do not admit simple normal forms. 

Commonly, GNN models (and in particular those based on the Message Passing paradigm \cite{mpnn}) organize computations in a \textit{synchronous} update scheme, i.e. all nodes update their state and exchange messages simultaneously. However, the nodes of the graphs we consider in this context have well-defined dependency relationships, hence following the approach adopted in \cite{dvae}, we process nodes sequentially, i.e. we establish a topological order on the nodes of the input graph (given by the depth-first traversal of the tree) and we perform computations following an \textit{asynchronous} scheme, by updating a node state only when all of its predecessors' states have been updated. Possibly, this order can be reversed (i.e. we consider the input DAG with edges in the opposite direction, adding a virtual root for the reversed graph connected to all the leaves of the original tree) and messages can be also exchanged in a bottom up scheme, originating what we refer to as a \textit{bidirectional} architecture. In a sense, when considering the bidirectional message passing scheme, the way in which computations are performed resembles the way in which one would evaluate the input formula itself.

\subsection{Encoding}\label{subsec:encoder}
The encoder is the network responsible for mapping the discrete input graph $G$ to a continuous latent representation $\textbf{z}$, and in this case consists of a GNN with a (possibly bidirectional) asynchronous update scheme, as described earlier in this Section. In our model, the encoder is formulated by adapting the Graph Attention Network (GAT, \cite{gat}) to the asynchronous message passing scheme of \cite{dvae}, we defer to Appendix \ref{subapp:gnn} for the mathematical details.

The initial nodes hidden states are the one-hot-encoding of their type (note that increasing the number of variables allowed increases the number of node types). Then, once the message phase terminates, all nodes states are computed, and we take as output of the encoder $out_e$ the state of the node without any successor (referred to as \textit{end node}; if no such node exists, a virtual one is added to the input graph). In case of bidirectional encoding, a concatenation of the state of the end nodes of both computation directions is taken. As approximate posterior $q(\cdot)$ of Equation \ref{eq:vae-loss} and \ref{eq:cvae-loss} we consider a Gaussian distribution, whose mean and variance are obtained by feeding $out_e$ to two MLPs. 

\subsection{Decoding}\label{subsec:decoder}
Given a real-valued latent vector $\textbf{z}$, the goal of the decoder is to convert it to a discrete graph $G$ representing the AST of a formula $\varphi$. The main requirement for the decoder is validity, i.e. it should produce syntactically correct formulae. Towards this objective we decided (ablations are discussed in Section \ref{subapp:ablations}) to design a decoder which implements syntactic rules architecturally, i.e. that does not allow the generation of invalid formulae. 
This is integrated in a node-wise generation procedure (derived from that of \cite{dvae}) which constructs one node $v_i$ at a time and updates its state $\textbf{h}_{v_i}$ as (denoting by $\mathcal{N}(v_i)$ the set of $v_i$'s predecessors and by $x_{v_i}$ the one-hot-encoding of its type):
\begin{equation}
    \begin{split}
    \textbf{h}_{v_i} & = GRU(x_{v_i}, \textbf{h}_{v_i}^{\text{agg}}) \\
    \textbf{h}_{v_i}^{\text{agg}} & = \sum_{v_j\in \mathcal{N}(v_i)} g(\textbf{h}_{v_j}) \odot m(\textbf{h}_{v_j})
    \end{split}
    \label{eq:dec-node-update}
\end{equation}
where $GRU$ is a Gated Recurrent Unit \cite{gru} and $\odot$ represents the gated sum between a mapping network $m$ and a gating network $g$, implemented as MLPs. 

Decoding thus results in the following top-down iterative algorithm:
\begin{enumerate}
    \item Given the initial latent vector $\textbf{z}$, the starting graph state $\textbf{h}_G = \textbf{h}_0$ is produced by feeding it to a MLP with softmax activation. By construction, the type of the first created node $v_0$ is \textit{start}, i.e. it is a synthetic starting node; 
    \item Each node $v_i$, with $1\leq i\leq \max_v$ (where $\max_v$ is an input parameter specifying the maximum number of nodes admitted during generation), is generated as:
    \begin{enumerate}
        \item the type of $v_i$ is inferred by first computing a distribution over all possible types using an MLP with softmax activation taking as input the current graph state $\textbf{h}_G$, then sampling its type from such distribution;
        \item the hidden state of $v_i$ is updated using Equation \ref{eq:dec-node-update};
        \item the graph state is updated as: $\textbf{h}_G = \textbf{h}_{v_i}$;
    \end{enumerate}
    \item Nodes are expanded following a depth-first traversal of the tree, taking into account the arity of the operators represented by node types, this prevents the violation of syntactic constraints; 
    \item The graph generation stops when the tree cannot grow further (i.e. all leaf nodes are variables), or when $i = \max_v$ (this latter case is the only possibility of having invalid formulae as output).
\end{enumerate}

In Section \ref{sec:propkernel} (and more in detail in Appendix \ref{app:kernel}) we describe an effective methodology to compute finite-dimensional semantic embeddings of propositional formulae. We remark that, in order to construct such representations for a formula $\varphi$, it is not necessary to explicitly know $\varphi$ itself, but only its valuations of a set of assignments $\{\tau_i\}_{i=1}^n \subseteq \mathcal{T}$. This makes it sensible to formulate a conditional variant of our model, i.e. a CVAE architecture using as semantic context vector the kernel embedding $\textbf{y}_{\varphi}$ of the input formula $\varphi$ (obtained after performing PCA dimensionality reduction), concatenated to the latent representation $\textbf{z}$ before starting the decoding computations.  

\subsection{Learning}\label{subsec:train}
Training is performed using teacher forcing, i.e. the decoder is fed with the ground truth graph it has to reconstruct during loss computation. At each iteration of the decoding algorithm of Section \ref{subsec:decoder}, we accumulate the negative log-likelihood by feeding the network with a vertex having the ground truth node type, so that the network is forced to stay close to true node sequences.

\section{Experiments}\label{sec:exp}
Experiments on a family of variations of our model pursue mainly two goals: measure VAE reconstruction and generative abilities and qualitatively evaluate the smoothness of the latent space. Hence, following \cite{dvae, grammar-vae}, we test:
\begin{itemize}
    \item Abilities of VAE models: we conduct standard experiments to check reconstruction accuracy, prior validity, uniqueness and novelty;
    \item Abilities of CVAE models: apart from reconstruction accuracy experiments as those of the previous point, we test the capability of our model of preserving semantic similarity in the latent space; 
    \item Latent space visualization: we qualitatively evaluate the ability of the latent space to capture characteristic structural features by visualizing it and interpolating it.  
\end{itemize}

\subsection{Experimental Setting}\label{subsec:setting}
We propose the following versions of our model (hereafter referred to as LogicGVAE): 
\begin{itemize}
    \item Syntax-only: given an input formula $\varphi$, it encodes its AST without any additional information, i.e. it has access only to the syntactic information of $\varphi$;
    \item Semantic-conditioned: given an input formula $\varphi$, we evaluate its semantic-context vector using the kernel of Section \ref{sec:propkernel} and perform conditional decoding as described at the end of Section \ref{subsec:decoder};
\end{itemize}

For both variants, we perform experiments on a set of different asynchronous encoding GNNs, namely GRU, Graph Convolutional Networks (GCN, \cite{gcn}) and GAT. Architectural and training details of all models are specified in Appendix \ref{app:more-exp}.

\subsection{Experimental Results}\label{subsec:results}

\subsubsection{VAE and CVAE abilities}\label{subsubsec:vae-abilities} 
This suite of experiments aims at evaluating: (a) accuracy, i.e. ability to perfectly reconstruct the input AST; (b) validity, i.e. ability to generate syntactically valid formulae; (c) uniqueness, i.e. proportion of distinct graphs out of valid generations; (d) novelty, i.e. proportion of generated graphs which are not in the training set. To evaluate accuracy, we encode each test graph $G$ to get mean and variance of the posterior approximation $q_{\phi}(\textbf{z}|G)$, then we sample $10$ latent vectors from such distribution and decode each $10$ times. Accuracy is defined as the percentage of these $100$ graphs which are identical to the input $G$. Validity is instead computed as the proportion of syntactically valid AST obtained by sampling $1000$ latent vectors  from the prior $p(\textbf{z})$ and decoding each $10$ times. For what concerns the conditional variant of our architecture, $q_{\phi}(\cdot)$ and $p_{\psi}(\cdot)$ are those of Equation \ref{eq:cvae-loss}.  

In the conditional setting, we test accuracy as described above, then we test how well the learned latent space preserves semantic-similarity by computing: (a) mean distance from the input context vector, i.e. given a semantic vector $\textbf{y}$ on which we condition the decoding, we sample $100$ vectors from the prior $\textbf{z}\sim p_{\psi}(\textbf{z}|\textbf{y})$ and decode each $10$ times, the most commonly decoded formula is kept for each $\textbf{z}$ and its semantic embedding computed and compared with ground truth $\textbf{y}$ using $L_2$ norm; (b) mean kernel similarity among formulae with same semantic vector $\textbf{y}$, i.e. following the procedure of the previous point, we compute the kernel among the most decoded formulae for each $\textbf{y}$. 

\begin{table*}
\centering
\captionsetup[table]{position=top}
\captionsetup[subtable]{position=top}
\caption{Results of VAE and CVAE abilities tests, percentages are averaged over $300$ test formulae with $5$ variables. The second line of the accuracy column for the GAT and GCN models represents the accuracy computed by considering only the most frequently decoded formula for each datapoint.}
\begin{subtable}{.45\linewidth}
\centering
\caption{Results of VAE abilities tests.}
\label{tab:vae-abilities}
\footnotesize
\begin{tabular}{lc|c|c|c}\toprule
   Encoder      & Acc. & Val. & Uniq. & Nov. \\\midrule
GAT  & \textbf{93.92} & 89.38 & \textbf{56.53} & \textbf{55.34}   \\
  & \textbf{95.42} &  & &  \\
 GCN & 91.27 & \textbf{98.78} & 29.91 & 28.32 \\
  & 93.56 & & & \\
GRU & 82.24 & 75.52 & 10.63 & 19.16 \\ \bottomrule
\end{tabular}
\end{subtable}
\hfill
\begin{subtable}{.45\linewidth}
\centering
\caption{Results of CVAE abilities tests.}
\label{tab:cvae-abilities}
\footnotesize
\begin{tabular}{lc|c|c|c}\toprule
 & Acc. & Val. & Sem.  & Ker.    \\
       Encoder  &  &  & Dist. & Value \\
   \midrule
GAT & \textbf{87.43} & \textbf{93.72} & \textbf{6.317} & \textbf{0.7985} \\
 & \textbf{92.45} & & & \\
 GCN & 85.35 & 89.14 & 6.808 & 0.6924     \\ 
 & 91.79 & & &  \\ \bottomrule
 \end{tabular}
 \end{subtable}
\end{table*}

Table \ref{tab:vae-abilities} reports results of the tests checking VAE abilities for different encoders (described in full details in Appendix \ref{app:more-exp}). The first aspect to notice is that a convolution-style GNN (either GCN or GAT) outperforms a recursive one (GRU) on all performance indices. This might be due to the fact that convolutions exploit more local substructure patterns and they leverage the hidden state of the node we are updating, instead of only those of its neighbors, differently from a GRU architecture. The reconstruction accuracy of both GCN and GAT is high and comparable, however when encoding with a GAT we record a higher ability of the latent space to capture structural features of data, as witnessed by a higher percentage of unique and novel decoded formulae. These results can be leveraged in a requirement optimization scenario, by mapping a given formula in its latent representation and optimize its structure in the continuous embedding space. 

Table \ref{tab:cvae-abilities} reports results of CVAE abilities tests (using as encoding networks only the most promising ones according to Table \ref{tab:vae-abilities}). If we take a pool of $5000$ random Boolean formulae with $5$ variables, which we use to generate the semantic context vectors for each test instance, they have an average distance between semantic embeddings of $18.7325$ and an average kernel similarity of $0.1695$. Since the semantic distances and the kernel values reported in Table \ref{tab:cvae-abilities} are much lower (resp. higher), we observe that conditioning on a semantic vector our input data actually maps them in a space where semantic similarity is to some extent preserved. 

It is worth noting that, although the reconstruction accuracy of the conditional architecture (Table \ref{tab:cvae-abilities}) is lower than that of the unconditional architecture (Table \ref{tab:vae-abilities}),  the former might be leveraged in  a requirement-mining scenario, where only valuations of a formula in a set of configurations are observed (thus allowing the construction of the semantic context vector) and an explicit formula optimising a fitness function has to be found (i.e. decoded). Essentially, in this case we need to use the CVAE as a generative model, to sample formulae which (approximately) have  the semantics specified by a given semantic context vector, obtained by solving an optimization problem in the semantic embedding space. This task cannot even be defined in a vanilla VAE scenario, where only syntactic information is exploited.
Moreover, the reconstruction accuracy increases significantly if we compute it considering only the most frequently decoded formula for each test point, as reported as second row of the accuracy column of Table \ref{tab:cvae-abilities}.

More experimental results and ablation studies are reported in Appendix \ref{app:more-exp}, we underline in particular the cruciality of the syntactic constraints in the decoder architecture (Section \ref{subsec:decoder}).
 
\subsubsection{Latent Space Visualization}\label{subsubsec:latent-visual}
\begin{figure}[t]
     \centering
     \includegraphics[width=0.95\textwidth, keepaspectratio]{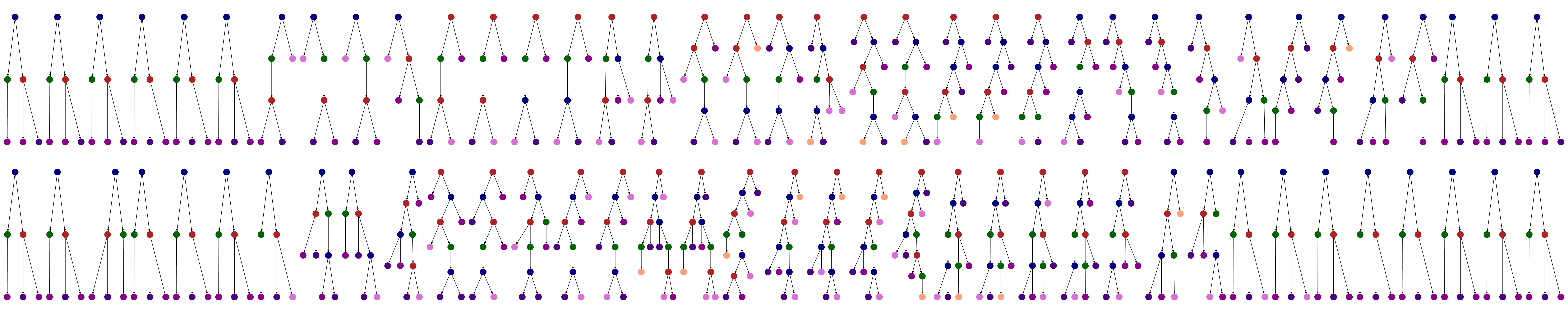}
     \caption{Slerp interpolation around a random formula of LogicVAE model with GAT (upper) and GCN (lower) encoder. Color of nodes encode their type.}
     \label{fig:slerp}
 \end{figure}

We use spherical interpolation (\textit{slerp}, \cite{latent-space}) around a given latent point to qualitatively evaluate the smoothness of the latent space. This briefly consists in interpolating points following a great circle of a hypersphere in a space having the same dimension of the latent space. Starting from a point $\textbf{z}_{\varphi}$ 
 encoding a formula $\varphi$, we follow a great sphere starting from $\textbf{z}_{\varphi}$ and pick $35$ evenly spaced points to decode. From Figure \ref{fig:slerp} we notice that the model using a GAT encoder produces a smoother latent space, as it changes less nodes at each step of the interpolation. 

\section{Conclusions and Future Work}\label{sec:conclusions}
In this work we propose LogicVAE, a configurable GVAE model built to learn a data-driven representation for propositional logic formulae. By leveraging an asynchronous message passing computational scheme in the encoder and ad-hoc syntactic constraints in the decoder, it is able to learn an expressive latent space. We provide a novel way to incorporate semantic information in the learning process, towards the goal of building an embedding space which is semantic-preserving. The main bottleneck of this approach is scalability to datasets with more than $5$ variables, a possible solution is a hierarchical approach, sketched in Appendix \ref{subapp:hierarchical}.

As already mentioned, results reported in this paper should be taken as a proof of concept towards extending the model to more complex formalisms, such as temporal logic, which come with the additional challenge of learning temporal and/or threshold parameters associated to operator (resp. variable) nodes, but which generally operates on few signals only (typically no more than $5$ \cite{ARCH20}). Moreover, inspired by AlphaCode \cite{alpha-code}, which defines a generative model for programs (another context in which the syntactic-semantic interplay has a fundamental role),  we plan to both incorporate a self-supervised pre-training stage in our encoder (adapting the masked language modelling loss used in AlphaCode), and to check the viability of using large language transformer-based models  \cite{transformer} to encode and decode logic formulae. 

\bibliographystyle{alpha}
\bibliography{biblio}

\clearpage
\appendix

\begin{center}
\textbf{\Large Appendix}
\end{center}

\section{Extended Background}\label{app:background}

\subsection{Propositional Logic}\label{subapp:logic}
Propositional logic is a formal language whose grammar recursively defines well-formed formulae by combining variables $\{x_i\}_{i=1}^n$ and logical operators $\neg$ (negation), $\wedge$ (conjunction) and  $\vee$ (disjunction) . More precisely, variables are valid formulae (also called atomic sentences), and given valid sentences $\alpha$ and $\beta$ it holds that $\neg \alpha$, $\alpha\wedge \beta$ and $\alpha\vee \beta$ are valid sentences as well. 

Semantics of propositional logic is defined starting from assignments $\tau$ of truth values to each propositional variable $x_i$ then recursively applying the following  rules to each sub-formula $\alpha$ (the valuation of $\alpha$ on $\tau$ is denoted by $\alpha(\tau)$): $\neg \alpha(\tau)=1$ if $\alpha(\tau)=-1$, $(\alpha \wedge \beta)(\tau)=1$ if both $\alpha(\tau)=1$ and $\beta(\tau)=1$, $(\alpha \vee \beta)(\tau)=1$ if either $\alpha(\tau)=1$ or $\beta(\tau)=1$.    

\begin{figure}[h!]
    \centering
    \includegraphics[width=0.3\textwidth]{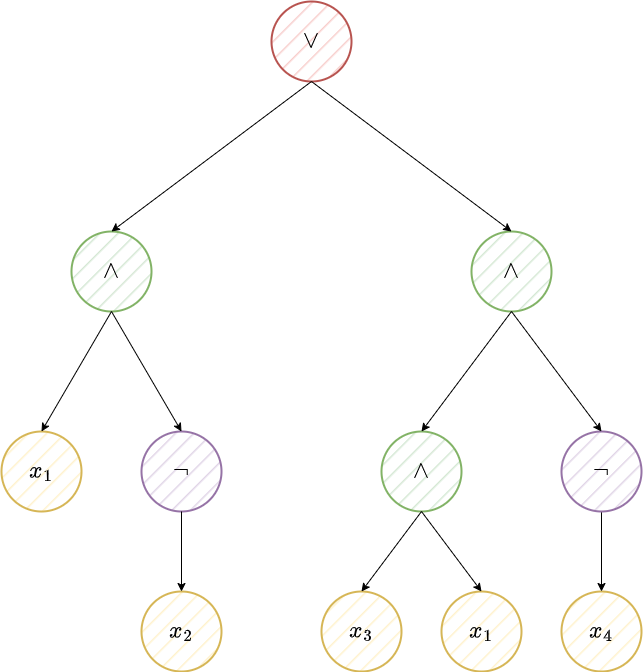}
    \caption{AST of the formula $(x1\wedge \neg x_2) \vee ((x_3\wedge x_1) \wedge \neg x_4)$. Color of the nodes encodes their type.}
    \label{fig:ast}
\end{figure}

As mentioned in Section \ref{sec:background}, every propositional formula $\varphi$ can be represented by its Abstract Syntax Tree (AST), as showed in Figure \ref{fig:ast}.

\subsection{Graph Neural Networks}\label{subapp:gnn}
Graph Neural Networks (GNNs, \cite{gnn-original}) are deep learning architectures that operate in the graph domain, i.e. they are built to solve graph related tasks in an end-to-end manner. 

Arguably the most common framework for GNN is the so called Message Passing Neural Network framework (MPNN, \cite{mpnn}), which abstracts many commonalities between existing GNN models. As the name suggests, the defining characteristic of MPNNs is that they use a form of neural message passing in which real-valued vector messages are exchanged between nodes (in particular between $1$-hop neighboring vertices), for a fixed number of iterations.

In detail, during each message passing iteration $t$ in a MPNN, the hidden representation $\textbf{h}_v^{(t)}$ of each node $v$ is updated according to the information $\textbf{m}_v^{(t)}$ aggregated from its neighborhood $\mathcal{N}(v)$ as:

\begin{equation}
\textbf{h}_v^{(t+1)}  = U_t(\textbf{h}_v^{(t)}, \sum_{w\in \mathcal{N}(v)} M_t (\textbf{h}_v^{(t)}, \textbf{h}_w^{(t)}))
\label{eq:mpnn_update}
\end{equation}
where $M_t$ is called message function and $U_t$ is called message update function: both are arbitrary differentiable functions (typically implemented as neural networks).

After running $T$ iterations of this message passing procedure, a readout phase is executed to compute the final embedding $\textbf{y}_v$ of each node $v$ as:

\begin{equation}
\textbf{y}_v = R(\{\textbf{h}_v^{(T)} | v\in G\})
\label{eq:mpnn_readout}
\end{equation}
where $R$ (the readout function) needs again to be a differentiable function, invariant to permutations of the node states.

Hence, the intuition behind MPNNs is that at each iteration every node aggregates information from its immediate graph neighbors, so as iterations progress each node embedding contains information from further reaches of the graph.  

A way to instantiate the message and update functions of Equation \ref{eq:mpnn_update} is the Graph Convolutional Network (GCN, \cite{gcn}) model, which generalizes the convolution operation to graph structured data by updating the representation $\textbf{h}_v$ of each node $v$ as:

\begin{equation}
    \textbf{h}_v^{(t+1)} = \sigma\bigg(\sum_{w\in \mathcal{N}(v)} \frac{1}{c_{vw}} h_w^{(t)} W^{(t+1)}\bigg)
    \label{eq:gcn}
\end{equation}
with $\sigma$ non-linear activation function and $c_{vw} = \{D^{\frac{1}{2}} (A + I)D^{\frac{1}{2}}\}_{vw}$ being $A$ and $D$ the adjacency and node-degree matrix of the input graph, respectively, and $I$ the identity matrix. 

Another possible computation scheme for the operations of Equation \ref{eq:mpnn_update} is the Graph Attention Network (GAT , \cite{gat}) which uses self-attention to learn relative weights between neighboring nodes. It consists in: 
\begin{equation}
\begin{split}
    \textbf{h}_v^{(t+1)} & = \sigma(\sum_{w\in \mathcal{N}(v)\cup v} \alpha_{vw}^{(t+1)} W^{(t+1)} \textbf{h}_w^{(t)}) \\
    \alpha_{vw}^{(t+1)} & = \text{softmax} (\text{LeakyReLU}(\textbf{a}^T[W^{(t+1)} \textbf{h}_v^{(t)} || W^{(t+1)} \textbf{h}_w^{(t)}]))
\end{split}
    \label{eq:gat}
\end{equation}
being $\sigma$ a non-linear activation function, $\textbf{a}$ a vector of learnable parameters and $||$ a node-wise operation like concatenation or sum. Possibly GAT uses a multi-head attention mechanism to increase the model capabilities, by replicating the operations of Equation \ref{eq:gat} multiple times independently and aggregating the results.

In its original version, the MPNN model is expected to exchange messages between nodes synchronously,  however when the input graph is directed it is sensible to follow a partial order among nodes, as observed in \cite{dvae}. In this \textit{asynchronous} message passing scheme, each iteration $t$ of Equation \ref{eq:mpnn_update} is performed following the topological order established for the nodes of the input graph: in this scheme  a node $v$ must wait for all of its predecessors' updates before computing its new hidden state $\textbf{h}_v^{(t)}$, i.e. it uses the current layer representation of its direct neighbors $\{\textbf{h}_w^{(t)}\}_{w\in \mathcal{N}(w)}$ before updating its state, instead of the one of the previous layer.

\section{Kernel Trick for Logic Formulae}\label{app:kernel}
\paragraph{A kernel function for Signal Temporal Logic} The kernel we start from when defining a semantic similarity measure for logic formulae is defined in \cite{stlkernel}. It is originally developed for Signal Temporal Logic (STL \cite{stl-original}), which is a linear-time temporal logic suitable to express properties over real-valued trajectories. A trajectory in this context is a function $\xi: I\rightarrow D$ where $I\subseteq \mathbb{R}_{\geq 0}$  is a time domain and $D\subseteq \mathbb{R}^n, n\in \mathbb{N}$ is a state space (we also denote by $\mathcal{T}$ the space of trajectories). A key characteristic of STL is that it can be given both a \textit{qualitative} (or Boolean) and a \textit{quantitative} notion of satisfaction (the latter called robustness). More in detail, given a STL formula $\varphi$, a trajectory $\xi$ and a time-point $t$, for the qualitative satisfaction we denote by $s(\varphi, \xi, t)=1$ (resp. $s(\varphi, \xi, t)=-1$) if $\xi$ at time $t$ satisfies (resp. unsatisfies) $\varphi$, while for the quantitative satisfaction we denote by $\rho(\varphi, \xi, t)\in \mathbb{R}$ the robustness of $\varphi$ for trajectory $\xi$, i.e. how robust is the satisfaction of $\varphi$ w.r.t. perturbations in the signal. It holds that $\rho(\varphi, \xi, t)>0 \rightarrow s(\varphi, \xi, t)=1$ and $\rho(\varphi, \xi, t)<0 \rightarrow s(\varphi, \xi, t)=-1$ (completeness property). 

The definition of quantitative satisfaction allows to consider STL predicates $\varphi$ as functionals mapping trajectories to their robustness, i.e. $\rho(\varphi, \cdot): \mathcal{T}\rightarrow \mathbb{R}$, hence a formula can be embedded into a (possibly infinite-dimensional) Hilbert space. In this space we can consider as `scalar product` between pairs of  formulae $\varphi, \psi$ the following:
\begin{equation}
    k(\varphi, \psi) = \int_{\xi\in\mathcal{T}} \rho(\varphi, \xi) \cdot \rho(\psi, \xi) \diff \mu_0 \xi 
    \label{eq:stl-kernel}
\end{equation}
being $\mu_0$ a probability measure over trajectories. It can be proved that $k(\cdot, \cdot)$ of Equation \ref{eq:stl-kernel} is a proper kernel function, and Probably Approximate Correct (PAC) bounds can be provided to give probabilistic bounds on the error committed when using the kernel for learning in formulae space. Experiments carried out in \cite{stlkernel}  confirms that the kernel of Equation \ref{eq:stl-kernel} efficiently captures semantic similarity between STL requirements. 

\paragraph{A kernel function for propositional logic} Building upon the idea of the STL kernel defined above, we define a similarity measure between propositional formulae as detailed in Section \ref{sec:propkernel}.

The Jaccard index (or similarity coefficient) measures the similarity between two sets $A$ and $B$ as $J(A, B) = \frac{|A\cap B|}{|A\cup B|}$ ; given two propositions $\varphi, \phi$ and the sets of their assignment valuations $X_{\varphi}, X_{\psi}$ resp. over variable configurations $\mathcal{T}$, the Jaccard index $J(\varphi, \psi) = \frac{|X_{\varphi}
\cap X_{\psi}|}{|X_{\varphi}\cup X_{\psi}|}$ denotes the proportion of configurations in which the two formulae agree, hence measuring semantic similarity between its inputs. 
On the other hand, the value of $k(\varphi, \psi)$ represents the normalized difference between the set of configurations $\tau\in\mathcal{T}$ on which $\varphi$ and $\psi$ agree and disagree, being $1$ in the case of semantically equivalent formulae. This quantity is positively correlated to the Jaccard index as $k(\varphi, \psi) = 2J(\varphi, \psi) - 1$, corroborating the claim that Equation \ref{eq:boolean-kernel} can be used as a measure of semantic similarity between formulae.  Putting all together, we get that the kernel is an interpretable measure of similarity between formulae, for example a kernel of $0.8$ can be understood as formulae agreeing on $ 90\%$ of configurations.

\begin{figure}[t]
    \centering
    \includegraphics[width=0.5\textwidth, keepaspectratio]{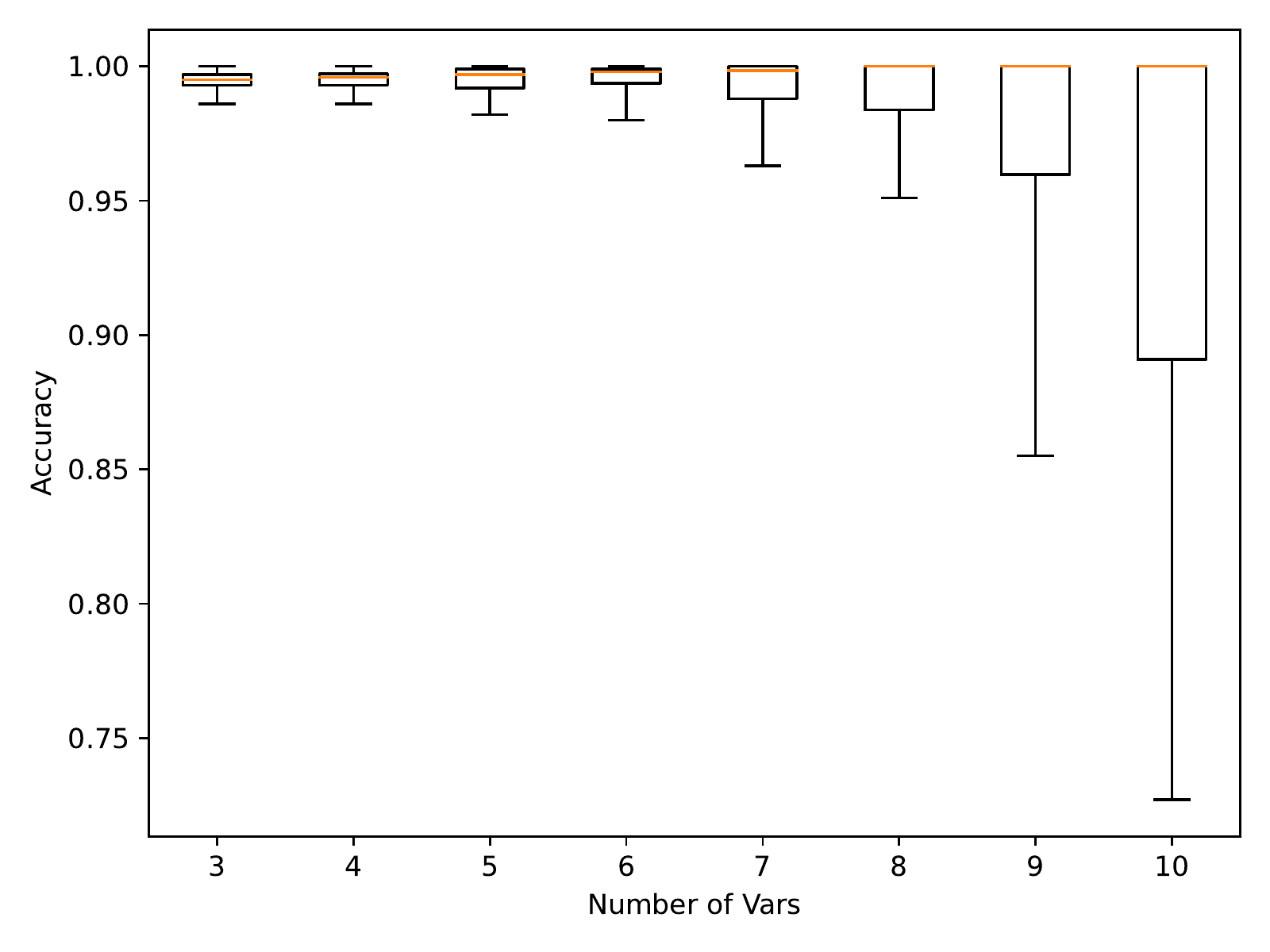}
    \caption{Accuracy of kernel classification varying the number of variables of training and test formulae, quantiles are averaged over $100$ experiments.}
    \label{fig:kernel-class}
\end{figure}

Moreover, we can give a deeper insight on the performance of the Boolean kernel in measuring semantic similarity between propositional formulae by showing the results of kernel classification on the satisfaction of assignments to formulae, reported in Figure \ref{fig:kernel-class}.

\begin{figure}
    \centering
    \includegraphics[width=0.5\textwidth, keepaspectratio]{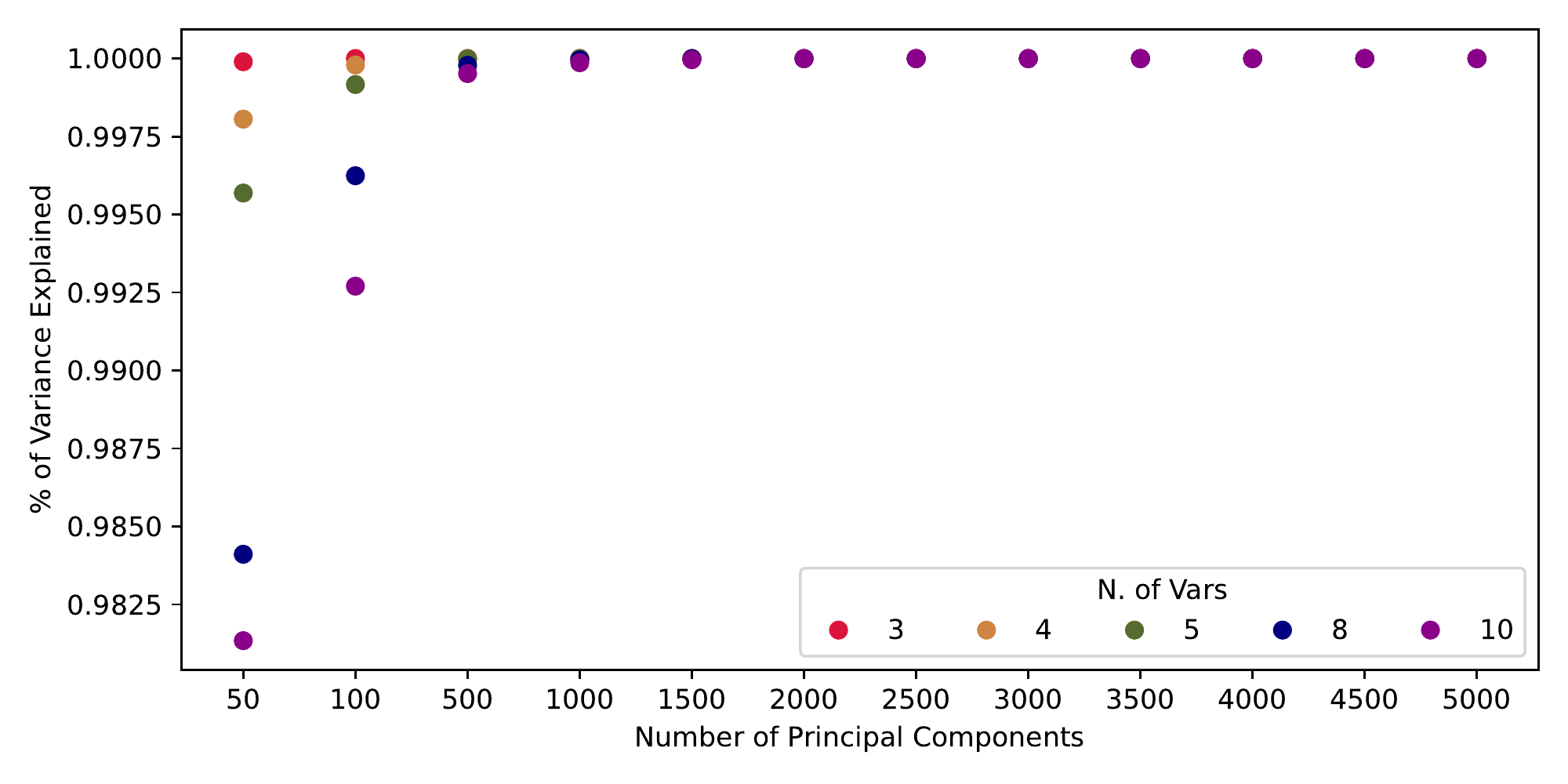}
    \caption{Proportion of explained variance against the number of principal components kept (kernel computed using $5000$ formulae), for different number of variables appearing in formulae.}
    \label{fig:kpca-variance}
\end{figure}

As the dimensionality of the embeddings computed by the kernel is equal to the number of training formulae used to evaluate the kernel itself (i.e. it is the vector of kernel evaluations according to Equation \ref{eq:boolean-kernel} of the input formula against all the formulae in the training set), we verify if it can be reduced without significant information loss by performing dimensionality reduction in the embedding space using Kernel Principal Component Analysis (kernel PCA). We answer this question affirmatively by showing that keeping $100$ components for formulae having at most $5$ variables and $500$ components for formulae with $6$ to $10$ variables retains much of the variance of the original dataset made of $5000$ formulae, as shown in Figure \ref{fig:kpca-variance}.  

\section{More Experimental Details}\label{app:exp}

\subsection{Dataset Generation Algorithm}\label{subapp:data}
In order to construct synthetic datasets of propositional formulae we adapt the algorithm defined in \cite{stlkernel} to our framework. It results in the recursive growing scheme for the syntax tree of each formula described in Algorithm \ref{alg:datagen}, where $p_{\text{leaf}}$ is the probability for a node to be a leaf node, $n$ is the max number of variable indexes allowed and $\textit{Cat}_{p_{\text{leaf}}}(\{\wedge, \vee, \neg, \text{VAR}\})$ denotes the categorical probability distribution assigning $p_{\text{leaf}}$ probability to the outcome $\text{VAR}$ (i.e. leaf node of the syntax tree) and $\frac{1-p_{\text{leaf}}}{3}$ to each operator node (i.e. internal nodes of the syntax tree).  

\begin{algorithm}
\caption{Syntax-tree random recursive growing scheme}\label{alg:datagen}
\begin{algorithmic}
\Require $p_{\text{leaf}}\in (0, 1)$, $n\in \mathbb{N}$  
\State $q \gets \emptyset$ \Comment{Queue of uncomplete nodes}
\State $\text{root} \gets \mathcal{U}(\{\wedge, \vee, \neg\})$ \Comment{Root is an operator node}
\State $q.\text{push}(\text{root})$
\While{$q \neq \emptyset$}
\State $N \gets q.\text{pop}()$ \Comment{Next node to expand} 
\If{$N\in \{\wedge, \vee\}$}
    \State $M, R \gets  \textit{Cat}_{p_{\text{leaf}}}(\{\wedge, \vee, \neg, \text{VAR}\}) \times \textit{Cat}_{p_{\text{leaf}}}(\{\wedge, \vee, \neg, \text{VAR}\})$
    \State $N.\text{left\_child} \gets M$
    \State $N.\text{right\_child} \gets R$ 
    \State $q.\text{push}(M)$
    \State $q.\text{push}(R)$
\ElsIf{$N$ is $\neg$}
    \State $M \gets \textit{Cat}_{p_{\text{leaf}}}(\{\wedge, \vee, \neg, \text{VAR}\})$
    \State $N.\text{child} \gets M$
\Else  \Comment{$N$ is a variable node}
    \State $N.\text{index} \gets \mathcal{U}(n)$ \Comment{Variable index is sampled randomly}
\EndIf
\EndWhile

\end{algorithmic}
\end{algorithm}

\subsection{More Training Details}\label{subapp:train}
All the models described in Section \ref{sec:exp} have been implemented in Python exploiting the PyTorch \cite{pytorch} library for leveraging GPU acceleration. 

Then, the loss function we optimize via mini-batch gradient descent is that of Equation \ref{eq:vae-loss} or its variant of Equation \ref{eq:cvae-loss} in the conditional version of our architecture. We allow the $KL$ term of such losses to be weighted by a tunable hyperparameter $\beta > 0$, i.e. using the loss of the so-called $\beta$-VAE \cite{beta-vae} model.  

Each architecture has been trained with Stochastic Gradient Descent (SGD) on a set of $4000$ different propositional formulae (generated as described in Section \ref{subapp:data}) divided in minibatches of size $32$  with an initial learning rate of $10^{-3}$,  using the Adam optimizer \cite{adam} and early stopping to regulate the number of training epochs (we put a validation step every $30$ epochs and stop the training if validation loss does not significantly improve with a patience of $3$ checkpoints; this resulted on having the non-conditional models trained for $\sim 300$ epochs and the conditional ones for $\sim 600$ epochs).  

Datasets for training and testing the models have been generated following algorithm \ref{alg:datagen}, fixing $p_{\text{leaf}}=0.4$ and varying $n\in \{3, 4, 5\}$. This resulted in sets having an average of  $[10.2955, 13.6632, 17.7475]$ nodes and average depth of $[5.4687, 8.5132, 9.4419]$ for increasing $n$. We remark that increasing the number of variable indexes allowed increases the number of node types the network has to learn (hence further increasing the complexity of the datasets).

We tested three different encoder architectures, each consisting on a GNN with the following distinctive features:
\begin{itemize}
    \item GRU, same as \cite{dvae}: the encoder is a MPNN where, using the notation of Equation \ref{eq:mpnn_update}, $M_t$ is a GRU and $U_t$ a gated sum (i.e. it follows the same update scheme of the decoder in Equation \ref{eq:dec-node-update});
    \item GCN: the encoder is a GCN (Equation \ref{eq:gcn}) with $2$ layers, unless differently specified;
    \item GAT: the encoder is a GAT with $3$ layers having $(3, 3, 4)$ heads resp., using sum as node-wise operation of Equation \ref{eq:gat} and concatenation as aggregator of single heads results in the internal layers and average in the final layer, unless differently specified.
\end{itemize}
All of them implement the asynchronous message passing scheme. For all the encoders,  we report results for the bidirectional setting, as justified by ablation studies in Appendix \ref{subapp:ablations}. 
To compute the mean and covariance of the approximate latent distribution $q_{\phi}(\cdot)$ we use two $1$-layer MLPs.

The hidden size of the models is set to $250$ and the latent size to $56$, these hyperparameters as well the ones listed above have been tuned with the hyperparameter optimization framework Optuna \cite{optuna}. As noted in \cite{dvae}, we also find that setting $\beta > 0.001$ as weight for the KL divergence in the loss function causes a significant drop in reconstruction accuracy. 

For what concerns the semantic context vector used in the CVAE version of our model, we computed it by kernel PCA as described in Section \ref{app:kernel}. Since our train and test datasets contain formulae having at most $5$ variables, we keep $100$ components, as justified by Figure \ref{fig:kpca-variance}.

\section{More Experimental Results}\label{app:more-exp}

In Tables \ref{tab:vae-abilities-app} and \ref{tab:cvae-abilities-app} we report more VAE and CVAE abilities results on different datasets of increasing complexity. We remark that GAT is the best performing encoder, both for the syntactic and the semantic-conditioned task. In particular in the conditional case it is able to better capture the multimodality of each $\textbf{y}$ w.r.t. to formulae $\varphi$ and give a better characterization of the semantics in the latent space. 

\begin{figure}
    \centering
    \includegraphics[width=0.75\textwidth, keepaspectratio]{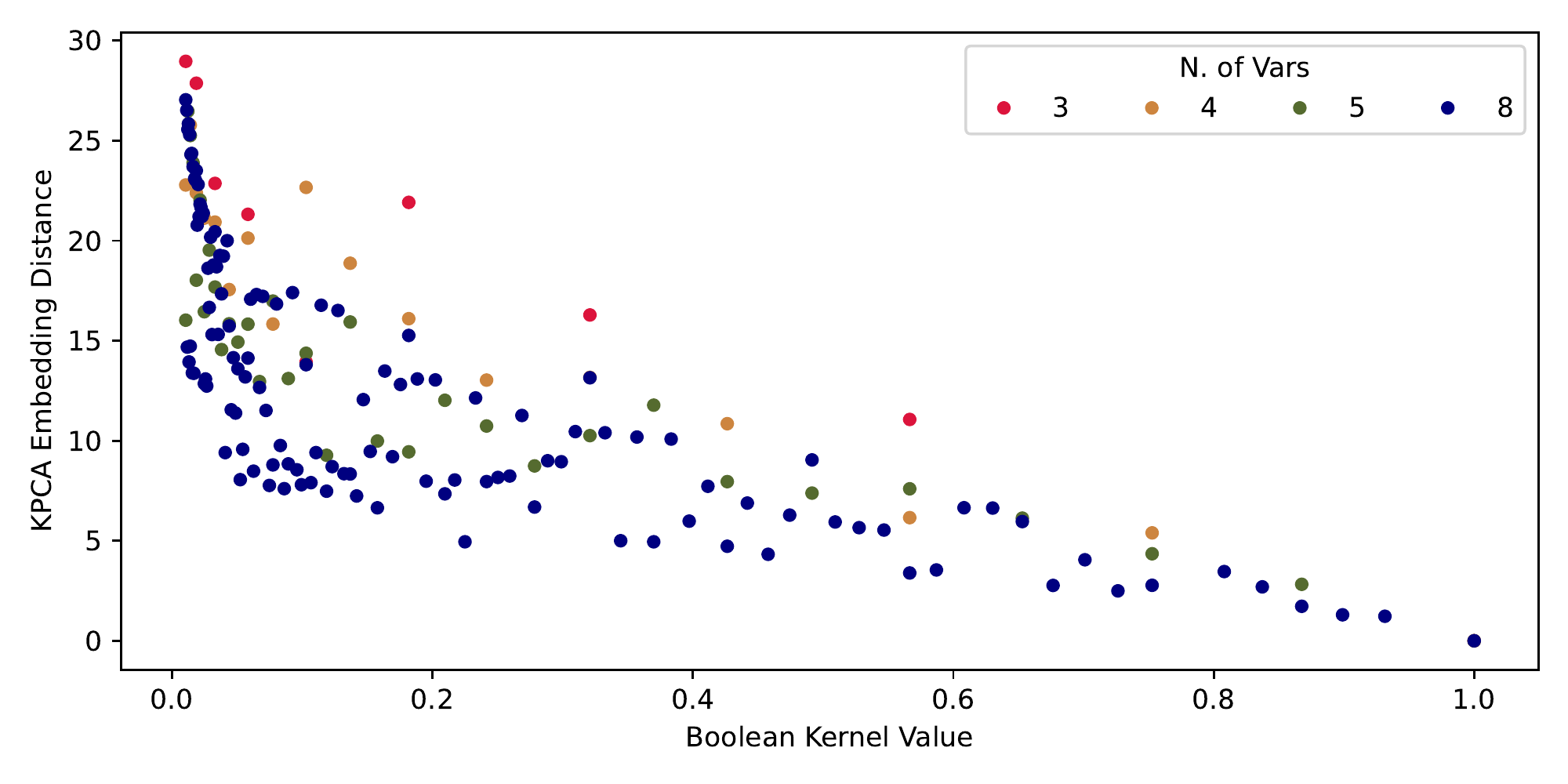}
    \caption{Correlation between kernel PCA embedding and kernel similarity between formulae, varying the number of variables for a pool of $500$ formuale.}
    \label{fig:kerndistance}
\end{figure}

Concerning the latter, Figure \ref{fig:kerndistance} gives an insight on the relation between kernel similarity value between a formulae $(\varphi, \psi)$ and distance between their kernel PCA embeddings. Being the kernel evaluated between two formulae an interpretable measure of their semantic similarity (more on this in Section \ref{app:kernel}), this reported inverse correlation justifies the metrics used for evaluating CVAE abilities. 

For what concerns the validity of formulae, we recall that the only case in which our decoding algorithm can produce invalid formulae is when it reaches the maximum number of nodes $\max_v$ allowed in the produced graphs, in our experiments set to $30$, as described in Section \ref{subsec:decoder}.

\begin{table}[t]
\centering
\caption{Results of VAE abilities tests, averaged over $300$ test formulae. The second line of the accuracy column for the GAT and GCN models represents the accuracy computed by considering only the most frequently decoded formula for each datapoint.}
\label{tab:vae-abilities-app}
\footnotesize
\begin{tabular}{lc|c|c|c|c|c|c|c|c|c|c|l}\toprule
& \multicolumn{4}{c}{$3$ var} & \multicolumn{4}{c}{$4$ var} & \multicolumn{4}{c}{$5$ var}\\\cmidrule(lr){2-5}\cmidrule(lr){6-9}\cmidrule(lr){10-13}
   Encoder      & Acc. & Val. & Uniq. & Nov.    & Acc. & Val. & Uniq. & Nov. & Acc. & Val. & Uniq. & Nov.\\\midrule
GAT  & \textbf{93.45}  & 93.42 & \textbf{58.09} & \textbf{54.24} & \textbf{93.96} & 93.70 & \textbf{54.13} & \textbf{53.59} & \textbf{93.92} & 89.38 & \textbf{56.53} & \textbf{55.34}  \\
& \textbf{96.43} &  & & & \textbf{97.32} & & & & \textbf{95.42} &  &  & \\
 GCN & 91.76 & \textbf{99.64} & 28.06 & 24.83 & 92.03 & \textbf{98.82} & 31.72 & 29.35 &  91.27 & \textbf{98.78} & 29.91 & 28.32 \\
& 94.23 & &  &  & 96.34 & &  &  & 93.56 & &  &  \\
GRU & 88.77 & 98.88 & 15.56 & 15.44 & 83.23 & 86.96 & 14.96 & 13.58 & 82.24 & 75.52 & 10.63 & 19.16 \\ \bottomrule
\end{tabular}
\end{table}

\begin{table}[t]
\centering
\caption{Results of CVAE abilities tests, percentages are averaged over $300$ test formulae. The second line of the accuracy column for the GAT and GCN models represents the accuracy computed by considering only the most frequently decoded formula for each datapoint.}
\label{tab:cvae-abilities-app}
\footnotesize
\begin{tabular}{lc|c|c|c|c|c|c|c|c|c|c|l}\toprule
& \multicolumn{4}{c}{$3$ var} & \multicolumn{4}{c}{$4$ var}& \multicolumn{4}{c}{$5$ var}\\\cmidrule(lr){2-5}\cmidrule(lr){6-9}\cmidrule(lr){10-13}
       & Acc. & Val. & Sem.  & Ker. & Acc. & Val. & Sem.  & Ker. & Acc. & Val. & Sem.  & Ker.   \\
  Encoder        &  &  & Dist. & Value &  &  & Dist. & Value &  &  &  Dist. & Value \\
   \midrule
GAT  & 87.75  & \textbf{98.59} & \textbf{8.819} & \textbf{0.5025} & 88.11 & 97.55 & \textbf{7.512} & \textbf{0.7692} & \textbf{87.43} & \textbf{93.72} & \textbf{6.317} & \textbf{0.7985} \\
& 93.23 & & & & \textbf{92.76} & &  &  & \textbf{92.45} & & &   \\
 GCN & \textbf{89.43} & 98.34 & 9.052 & 0.4671 & \textbf{88.87} & \textbf{99.55} & 8.447 & 0.6115 & 85.35 & 89.14 & 6.808 & 0.6924   \\ 
 & \textbf{94.97} & & & & 92.45 & &  &  & 91.79 & & &   \\\bottomrule
\end{tabular}
\end{table}

For what concerns the accuracy of reconstructed formulae, we also notice that it increases if we consider, for each of the test formulae, only the most common decoded one (which is a more realistic scenario). Accuracy in this scenario for GAT and GCN encoders is reported as second line of the corresponding model in Tables  \ref{tab:vae-abilities-app} and \ref{tab:cvae-abilities-app} for the conditioned and unconditioned case, respectively.

In Figure \ref{fig:hist-cvae} we show the distributions of kernel values and semantic distances computed as described earlier for LogicVAE with both GAT and GCN encoder, when trained on a dataset of $5$ variables. Moreover, in Table \ref{tab:cvae-abilities-quantiles} we report their first, second and third quantiles.

\begin{table*}[t]
\centering
\begin{minipage}{.55\linewidth}
\centering
    \captionof{figure}{Distribution of kernel value (left) and semantic \\ distance (right) for GAT and GCN models, \\ on a dataset of $5$ variables.}
    \label{fig:hist-cvae}
    \includegraphics[width=\textwidth]{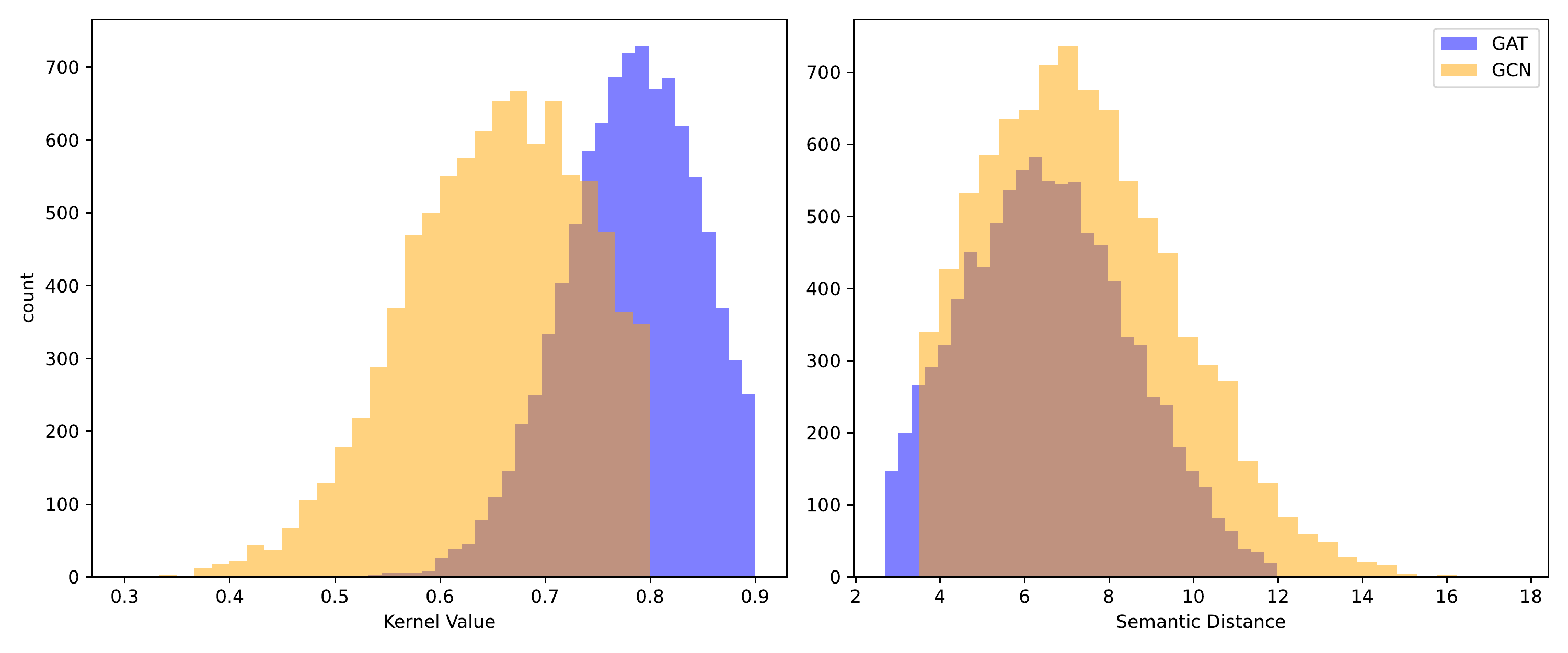}
\end{minipage}
\hfill
\begin{minipage}{.4\linewidth}
\centering
\caption{Quantiles of the distribution of kernel value \\ and semantic distance for GAT and \\ GCN models, on a dataset of $5$ variables.}
\label{tab:cvae-abilities-quantiles}
\footnotesize
\begin{tabular}{lc|c|c}\toprule
& \multicolumn{3}{c}{Kernel Values} \\ \midrule
 & Q1 & Median & Q3    \\
   \midrule
GAT & 0.7395 & 0.7843 & 0.8276  \\
 GCN & 0.5932 & 0.6578 &  0.7169 \\ \midrule
 
& \multicolumn{3}{c}{Semantic Distances} \\ \midrule
 & Q1 & Median & Q3    \\
   \midrule
GAT & 5.097 &  6.455 & 7.887  \\
 GCN & 5.658 &  7.166 &  8.846 \\ \bottomrule
 \end{tabular}
\end{minipage}
\end{table*}
\subsection{Ablations}\label{subapp:ablations}

In Table \ref{tab:ablations} we report the results of several ablation studies we perform on LogicVAE. They consists in the following:

\begin{enumerate}
    \item Non-constrained decoding: this represents the most important design choice. We kept the encoding and objective function as described earlier, but we remove from the decoder the constraints imposing to follow logic syntactic rules (i.e. the ones described in Section \ref{subsec:decoder}). This leads to a dramatic decrease in reconstruction accuracy, highlighting the fundamental importance of this inductive bias. 
    \item Unidirectional message passing: messages are exchanged only following the original direction of the edges of the input DAG (i.e. without reverting them). For both GCN and DAG encoders, this decreases the reconstruction accuracy, but not dramatically. 
    \item Number of convolutional/attentional layers: we get the highest accuracy with $2$-layer GCN encoder and $3$-layer GAT encoder.
\end{enumerate}

\begin{table}[t]
\centering
\caption{VAE accuracy on several ablations of the model, averaged over $300$ test formulae with $5$ variables.}
\label{tab:ablations}
\footnotesize
\begin{tabular}{lc|c|c}\toprule
        & Non-constrained & Unidirectional & \# layers  \\
   Encoder  & decoding & & $1$/$2$/$3$/$4$/$5$\\
   \midrule
GAT  & 32.53 & 89.66 & 81.38/89.21/\textbf{93.92} /91.37/87.57  \\
 GCN & 25.64 & 90.58 & 89.18/\textbf{91.27}/90.53/87.31/85.29   \\ \bottomrule
\end{tabular}
\end{table}

\subsection{Hierarchical Learning of Logic Formulae}\label{subapp:hierarchical}
While performing experiments, we noticed that LogicVAE struggled in learning variable indexes, i.e. its reconstruction accuracy increases (and the architectures converge faster) if we provide as input a simplified AST for each input formula $\varphi$. This graph is constructed by removing variable indexes from the input described in Figure \ref{fig:ast} and Section \ref{sec:method}. 
 The idea is then to build a hierarchical learning model, which first learns to reconstruct the simplified tree using the syntactic version of LogicVAE,  then recovers variable indexes by a GNN architecture. 
 We instantiated this latter model using a MPNN architecture, akin to that described in Section \ref{subapp:train}, with node-level readout function on the leaves. We trained the model to minimize a loss involving both a classification term (namely cross-entropy loss over $n$ classes, being $n$ the maximum number of variable indexes allowed) and a semantic term measuring the square loss between the semantic vector $\textbf{y}_{\varphi}$ of the ground-truth formula $\varphi$ (as described in Section \ref{subapp:train}) and the one of the reconstructed formula $\hat{\varphi}$, denoted by $\hat{\textbf{y}}_{\hat{\varphi}}$ at each step of the training algorithm. Formally we minimize the following (for a formula $\varphi$ having $M$ leaves over $n$ possible indexes):
 \begin{equation}
     \mathcal{L}(\varphi, \hat{\varphi}) = -\frac{1}{M}\sum_{v=1}^M(\sum_{i=1}^n y_{vi} \log \hat{y}_{vi}) + \lambda ||\textbf{y}_{\varphi} - \hat{\textbf{y}}_{\hat{\varphi}}||_2
     \label{eq:hier-loss}
 \end{equation}
 
where $y_{vi}, \hat{y}_{vi}$ represent the true (resp. reconstructed) probability of leaf $v$ to have index $i$, and $\lambda\in \mathbb{R}_{\geq 0}$ is a tunable parameter  weighting the semantic term of the loss w.r.t. the classification one. 
Interestingly, we found that the loss of Equation \ref{eq:hier-loss} reaches a plateau in correspondence of low values of the semantic term. Indeed, we verified that the true formula $\varphi$ and the reconstructed one $\hat{\varphi}$ evaluated in correspondence of the learning plateau differ on the $[10.34\%, 12.29\%, 12.48\%, 16.85\%]$ of possible configurations with $n\in [3, 4, 5, 10]$ (results with $\lambda=0.7$ found with hyperparameter optimization,  removing either the classification or the semantic term in the loss does not ameliorate the results). Hence, to some extent, a node-level MPNN learns to focus on semantically relevant parts of the input formula. We consider this results interesting towards a better understanding of the learning dynamics of GNN architectures in the context of learning logic formulae and we plan to investigate it deeper in the future. 

\end{document}